# Bombus Species Image Classification

George LaVezzi
*Department of Computer Science*
*Kansas State University*
Manhattan, KS, USA
gblavezz1@ksu.edu

Venkat Margapuri
*Department of Computer Science*
*Kansas State University*
Manhattan, KS, USA
marven@ksu.edu

Robert Stewart
*Department of Computer Science*
*Kansas State University*
Manhattan, KS, USA
rstewar2@ksu.edu

Dan Wagner
*Department of Computer Science*
*Kansas State University*
Manhattan, KS, USA
danwagner@ksu.edu

*Abstract*—Entomologists, Ecologists and others struggle to rapidly and accurately identify the species of bumble bees they encounter in their field work and research. The current process requires the bees to be mounted, then physically shipped to a taxonomic expert for proper categorization. We investigated whether an image classification system derived from transfer learning can do this task. We used Google's Inception, Oxford's VGG16 and VGG19 and Microsoft's ResNet 50. We found Inception and VGG classifiers were able to make some progress at identifying bumble bee species from the available data, whereas ResNet was not. Individual classifiers achieved accuracies of up to 23% for single species identification and 44% "top-3" labels, where a composite model performed better, 27% and 50%. We feel the performance was most hampered by our limited data set of 5,000-plus labeled images of 29 species, with individual species represented by 59 -315 images.

## I. INTRODUCTION

Dr. Brian Spiesman, from Kansas State University's College of Agriculture's Entomology Department, has identified a need for the rapid, accurate identification of bumble bees by species from images taken in the field by researchers. The current identification process involves capturing the bees, returning from the field, mounting the bees on pin boards, then shipping them to taxonomic experts for proper identification. This is both an expensive and time-consuming process, often requiring months from bee collection to proper identification; thus delaying the pace at which research can be conducted. A trained classifier, particularly one working from images in the wild (as opposed to dried, pinned and mounted) which can properly identify bumble bee species from images would be of tremendous help.

Contemporaneously, several pre-trained convolutional neural networks are available for transfer learning image classification tasks, such as Google's Inception, Oxford's VGG16 and VGG19, and Microsoft's ResNet 50. This offers the opportunity to compare their performance on the bumble bee task and opens the possibility of a composite model solution. Several of these models are implemented in the TensorFlow machine learning framework, which were to conduct this project.

## II. RELATED WORK

### A. Brief History of Image Classification

In the 1960s, Papert is credited with some of the earliest work in this area where image recognition and characterization are based on distinct feature identification (edges, textures, curve etc.) [1]. Techniques to identify and classify these features continued apace but were hampered by limited computational power and memory. In the 1980s, several algorithms were introduced (e.g. Canny Edge Detection) to improve this feature detection [2]. Deep learning techniques began to make their presence felt for feature extraction and pattern recognition in the 2000s with the advancement in processing power and memory capacity [3]. A CNN image segmentation won its first challenge in 2012 and dominated the field for several years thereafter [4].

### B. CNN for Image Classification

Image classification identifies the presence of an item of interest (a member of a class) in the picture in question. A classifier which recognizes both cats and dogs may classify an image with both a cat and a dog as one or the other based on some degree of "cat-ness" (or "dog-ness") computed by the network; this calculation may not have an exact human understandable analog.

A CNN is composed of two or more connected layers of neurons. At least one of the layers is convolutional, using a "window" (receptive field) to map a set of inputs, through the convolution operation to the neurons in the receiving layer. A given neuron's output is then determined by its convoluted input, a weighting value, bias value and activation function. Thus, a node's (neuron's) behavior looks akin to Figure 1, where the x_# are the convolutional results of the previous layer (or input).

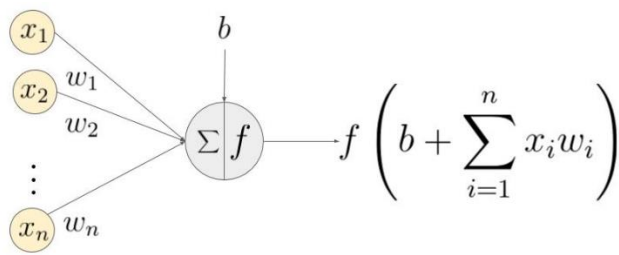

**Figure 1.** Single Neuron Activation [1]

The network itself may be like Figure 2, but with a different number of layers and without picturing the convolutional functions between layers.

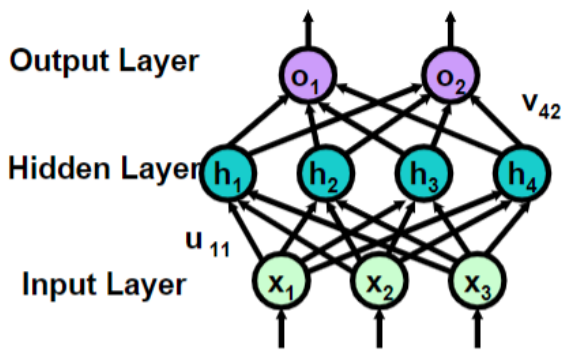

**Figure 2**. Representative CNN Layers [5]

The activation function is typically a differentiable non-linear function, such as the sigmoid or a rectifier liner unit (ReLU). The output layers are typically the class-labels themselves, with class selection based on a maximum or one-hot selection.

A CNN will have one or more convolution layers which look at collections of outputs from the previous layer (inputs), like the values of all the pixels adjacent to the pixel of interest and convolves (combines/filters) them. Such a convolution may filter, pool, etc. the incoming information as well as change its dimensionality. There can also be skip layers, upscaling layers, etc. to provide the "structure" or "encourage", or if you will, the abstractions that are appropriate to the classification task. The selection and ordering of layers appears to be based as much on empirical experience as theoretical footings.

Early layers (near the input layer) detect feature-analogs such as edges. Mid-layers are analogous to more complex features, such as color-histograms. Later layers (near the output) recognize objects. However, none of these layers necessarily has a human cognitive analog.

Krizhevsky, Sutskever, and Hinton used a CNN to classify over a million images in 2010. CNNs' capacities can be controlled by varying their depth and breadth, and they tend to make strong and mostly correct assumptions about the nature of images (namely, stationarity of statistics and locality of pixel dependencies). Figure 3 shows a portion (1 GPU's worth) of the CNN used by [6].

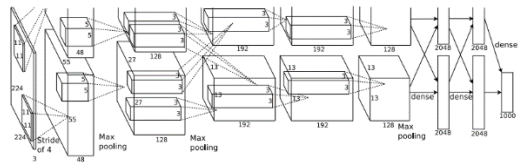

**Figure 3.** CNN used for image Classification

### C. CNNs and Transfer Learning

The types of features learned by early layers of a CNN trained on image data tend to have generalizable characteristics, whereas the latter layers tend to be more specific to the actual objects being classified [8]. This enables one to "graft" a pre-trained set of initial layers onto a blank or to-be-trained set of late layers to speed the training of the complete CNN. This approach is known as transfer learning—the features the "early layers" have been trained on are transferred to the new problem set.

It has become common practice to use generalized, pre-trained early feature detection layers, trained on hundreds of thousands to over a million images, connected to target network for "top-off" or "customization" training. This enables the target network to be trained more quickly with fewer images.

### D. Bee Classification

We encountered only two previous attempts at bee-image classification: Dr Spiesman's unpublished work using just the mounted and pinned forewings of bees (which resulted in 89% single species accuracy); and, a DrivenData hosted a crowd sourced competition to classify bees by genus in 2015. A solution for DataDriven's challenge using Google's Inception achieved a 99% AUC score on images of bees taken in the wild [14].

### III. PROBLEM STATEMENT

Identification of bumble bee species is difficult, requiring collection of live specimens in the wild. These specimens are mounted and physically shipped to a taxonomic expert for correct species categorization. This process takes a large amount of time from both the collectors and the expert; and is expensive. We propose developing an image based classifier whose goal is to accurately identify bumble bee species in an efficient manner to expedite the procedure.

### IV. TECHNICAL APPROACH

Using data provided by Dr. Spiesman, we developed a script which produced standardized training, augmented training, and test datasets for use in training the selected pre-trained classification engines, informed by several [9,10,11]. Each researcher used these sets to train their selected model and pursue some level of parameter fine tuning in a TensorFlow 2.0 framework. We will also build composite models.

We then characterized the performance of our individual and composite models with respect to single species accuracy and 3-subspecies grouping accuracy. As we are interested in bumble bee species accuracy, the test data does not contain non-bumble bee images; we did not want to skew the accuracy in case we achieved good performance in differentiating by genus (as was obtained by DataDriven) but poor performance in species differentiation.

We chose the VGG19, VGG16, ResNet50 and InceptionV3 pre-trained image classifiers for this experiment. All of these were obtained from TensorFlow-hub. The general approach was to vary the number of post-retrained additional hidden layers, the number of nodes in each added layer, learning rate, drop out, batch normalization and other techniques to control validation over fitting

## V. Experimental Setup

The bumble bee data for this experiment provided by Dr. Spiesman consists of over 5,000 images classified into 29 species. An additional "classification" of non-bumble bees, consisting of roughly 200 hundred labeled honey bee images from Kagel [12] was added so the data set would have both positive and negative examples to aid in learning generalizations. The images are predominately of bees in the wild and therefore contain random backgrounds, bee orientations. Some or most of the bee is often obscured. Additionally, the images lack standard size and resolution and are not evenly distributed by class.

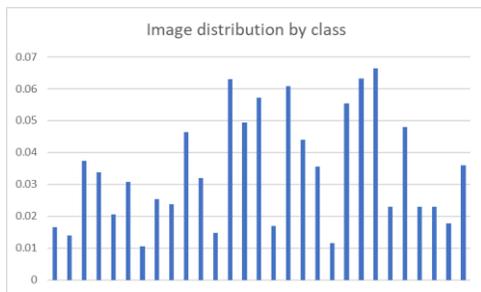

**Figure 4.** Bumble Bee Species Distribution

We created three standardized data sets: training, testing, augmented training. Image augmentation was performed by

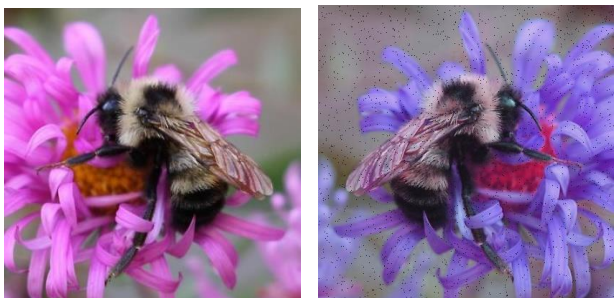

using a random combination of: rotation, contrast manipulation,

**Figure 5.** Image and Augmented Image

salt and peppering and adding obscuring blocks (randomly "zeroing out" a box of pixels in the image.) Roughly 25% of the images were augmented in the augmented training set. The segmentation of training data into training and validation was left to each researcher. Models were trained on both training sets for comparisons.

### A. VGG 19

The VGG19 [13] model was imported from Keras' built-in models with weights from the ImageNet dataset. The network was trained on a 16GB Intel I-7 hexacore CPU node with an integrated graphics card: all training was run on the processor rather than the GPU. Each layer was copied from the existing model into a new model except for the final three fully connected layers, including the categorization/output layer: this layer was modified to reflect our reduced number of categories (i.e. 1000 reduced to 30). The two FC layers before the categorization layer had their node size reduced due to learning stagnation on early iterations: this value worked well and did not improve as the hidden layer size was decreased further. Learning rate was experimentally altered and ranged from 0.00001 to 0.001 using a batch size of 64. Dropout was added between each FC layer to help with any overfitting and had a probability of 0.5.

The dataset that was used for training was split with 85% training and 15% validation data. These were fed to the model as generators; the training generator was shuffled on each set of training (5, 10, and 10 epochs) while the validation set remained constant for a consistent point of comparison. The model trained for up to 25 epochs with tweaking to prevent overfitting. Iterations of the model were run on both the normal and augmented datasets, with the former exhibiting better performance and faster learning than the latter. In general, the augmented dataset did not help: with the different resolutions and orientations of bees in the images, the dataset already exhibited a degree of augmentation.

### B. VGG16

The VGG16 model was imported from the canned architectures provided by Keras as part of its Applications module. The imported model comes with pre-trained weights from the ImageNet dataset. The model was trained in a CPU and a GPU environment where the CPU environment was an Intel i7 CPU running Windows 10 and the GPU environment was an Intel Core i7-6700k quad core processor, with an 8GB RTX 2080 graphics card, running Ubuntu 18.04.02 and TensorFlow 2.0 (nightly version).

VGG16 requires the images to be of size 224 x 224 x 3 pixels (pixel width x pixel height x RGB channels). Several architectures of the model with varying hyperparameter values were evaluated and are as shown in table 1.

**Table 1.** Hyperparameters for VGG16

| Hyper Parameter/Structure | Experimental Values |
|---|---|
| Number hidden trainable layers | 1 – 3 |
| Nodes per trained layers | 64 – 2048 |
| Optimizers | Adam, SGD |
| Learning Rate | 0.01 – 0.00001 |
| Learning Rate Decay | Yes |
| Drop out (b/w each FC layer) | 0.0 - 0.5 |

A Sequential model was built using each of the layers of the imported model with the exception of the final layer. In addition, between 1 and 3 fully connected layers were added to the models. None of the pre-trained layers were trained. Instead, only the newly added fully connected layers were trained.

Of the data in the training dataset, 80% of the data was used for training and 20% of the data was used for validation. All of the experimented architectures were trained for about 20 - 25 epochs after which the models started to overfit. Applying Learning Rate Decay did not help alleviate the problem. The use of the augmented dataset on the model only resulted in mediocre results.

*C. RESNET 50*

ResNet50 was imported from Tensorflow 2.0. The model was trained with an Intel Core i7-6700k quad core processor, 8GB RTX-2080, running on Ubuntu 18.04.02 and the GPU-enabled Tensorflow 2.0 (nightly version). ResNet50 requires all input images to be of size 224 x 224 x 3 (pixel width, pixel height, RGB values).

Throughout the experiment, we attempted numerous variations to hyperparameter values (e.g. learning rates, number of fully-connected layers, etc.), weight initialization, and batch amounts. In addition to varying hyperparameters, we also attempted different optimizers, namely Adam and Standard Gradient Descent (SGD). For SGD, we also tested with weight-decay and momentum both enabled and disabled.

**Table 2.** Hyperparameters for ResNet50

| Hyper Parameter/Structure | Value Range |
|---|---|
| Number hidden trainable layers before output layer | 0 – 3 |
| Nodes per trained layers | 64 – 2048 |
| Learning Rate | 0.001 – 0.000001 |
| Drop out | 0.0 -0.7 |
| Normalization | On, Off |
| Weight initialization | Imagenet, random |
| Epochs | 5-20 |
| Weight Decay | 0.01 |
| Momentum | 0.9 |

Ultimately, for the composite model, we decided to use zero additional hidden layers and instead have a GlobalAveragePooling2D layer before the final output layer.

The output layer consists of 30 nodes, representing the 30 distinct classes. Our final ResNet model had randomly initialized weights, used the Adam optimizer, with a learning rate of 5e-4, categorical cross-entropy as the loss function, and softmax as the activation function. We used an 80/20 training-validation split, with a batch size of 64 images, and trained for 15 epochs. Dropout, weight-decay, and momentum were not used for this model. The inspiration for these models can be found in [18] and [19].

*D. InceptionV3*

The Inception based classifier was trained on a 20 GB intel I-7 quad core with a 6GB GTX-1060 GPU, running Windows 10 and GPU enabled TensorFlow 2.0. InceptionV3 requires all images to be 299 x 299 pixels in size and native TensorFlow sizing functions were used to shrink/stretch each image as the data was loaded. The following structures and hyper parameters were varied in an effort to fine tune the model.

**Table 3.** Hyperparameters for InceptionV3

| Hyper Parameter/Structure | Value Range |
|---|---|
| Number hidden trainable layers | 1 – 3 |
| Nodes per trained layers | 128 – 2048 |
| Learning Rate | 0.001 – 0.000001 |
| Drop out | 0.0 -0.75 |
| Normalization | Attempted – did not help |

An 85/15 train validation split was used with 10-25 epochs of training being normal for each model. The limited size of the GPU memory forced batch sizes of less than 16 (12 was used). The models tended to badly over fit, even when drop out is used. Learning rate decay did not help when validation loss plateaued. Better results were obtained against both validation and test data sets when the model was trained with the normal (un augmented) data. The software for this model was strongly influenced by the TensorFlow Hub Authors [17].

*E. Composite Model*

We combined various combinations of the best trained modes into a composite model, by summing their softmax outputs and selecting the largest resultant values. Different combinations of "best model" were tried to see if a such a simple composite model can improve performance.

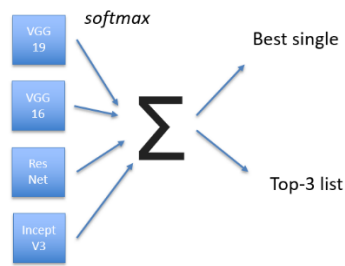

**Figure 6.** Conceptual Composite Model

## VI. RESULTS

Most of the models performed better on the normal (un-augmented) data training set. We hypothesize that the training set contains sufficient "noise", with its different orientations, resolutions and sizes, that good generalization is obtain without the need of "fuzzing" the images.

### A. Best Individual Models

*1) VGG19*

In the case of VGG19, the model trained the best with the two FC layers having 2048 nodes each, dropout 0.5, learning rate decay starting at 0.00001 and decaying by 0.96 every 100 epochs. ADAM was used with the decay rate and error was calculated via categorical cross entropy. Training stopped after 10 epochs on the final model due to consistent overfitting.

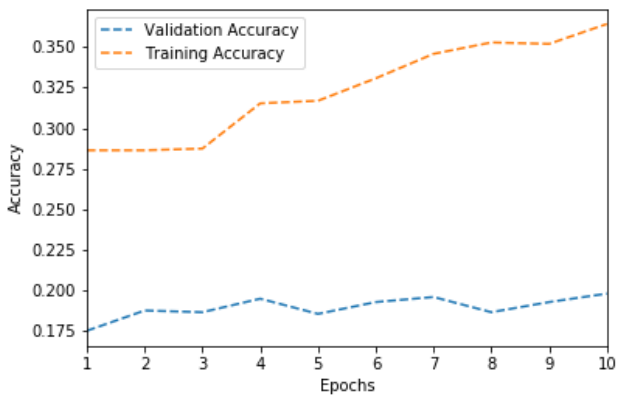

**Figure 7.** VGG19 Accuracy Plot

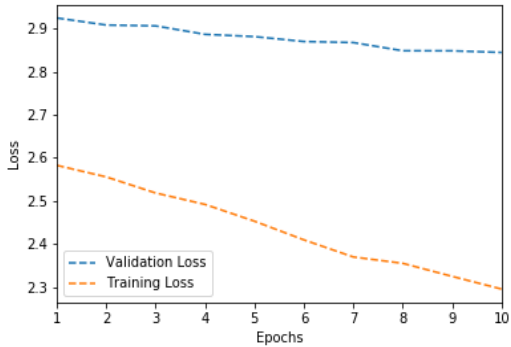

**Figure 8.** VGG19 Loss Plot

*2) VGG16*

The best VGG16 architecture had three additional fully connected layers each with 2048 nodes, a dropout of 0.3 between each of the fully connected layers, an optimizer of ADAM and a learning rate of 0.0001. 20 epochs of training was performed on the model before the model began to show signs of overfitting. In the following plots, the blue lines represent training data and the orange lines, validation.

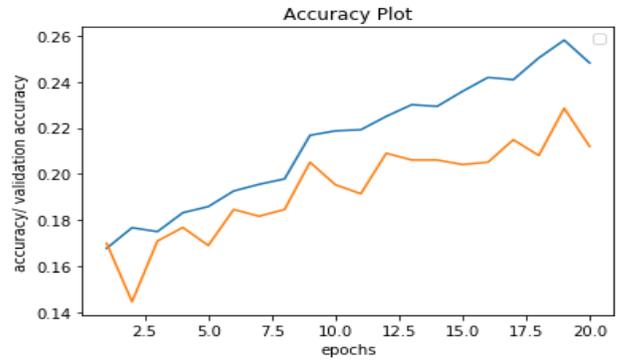

**Figure 9.** VGG16 Accuracy Plot

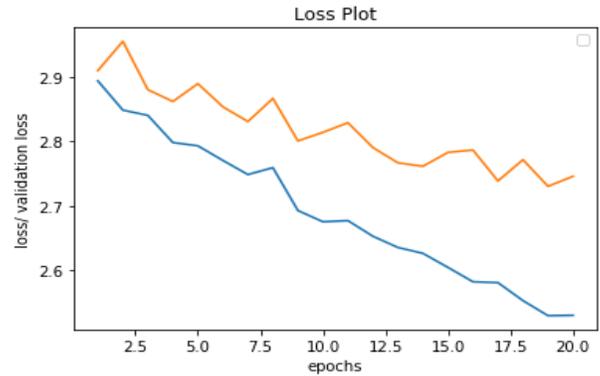

**Figure 10.** VGG16 Loss Plot

*3) ResNet 50*

ResNet50's best accuracy plots, figures 11 and 12, use blue for training data and orange for validation.. As can be seen, this is a clear indication of overfitting.

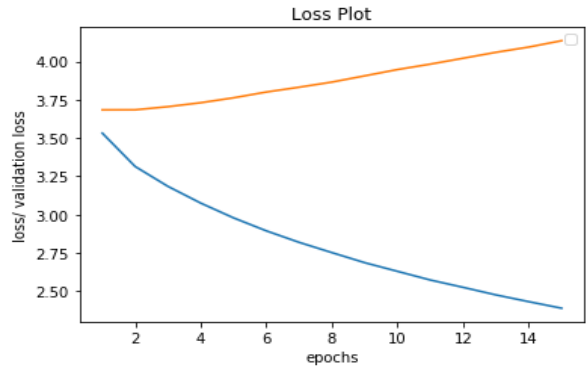

**Figure 11.** ResNet50 Loss Plot

Furthermore, the model's accuracy on the validation set never exceeded 3.33%. All hyperparameter tuning turned out to equally poor.

This was the principle reason for performing so many variations in the hyperparameters. Even with regularization methods in place and a small learning rate, the models never seemed to break out of the local minima it reached. We also attempted smaller blocks, as seen in [19], to reduce the possibility of over-relying on pre-learned features. The end result still did not change. We also froze and unfroze layers to determine if training from scratch would give better results. Still, the 3.33% validation accuracy remained unmoved.

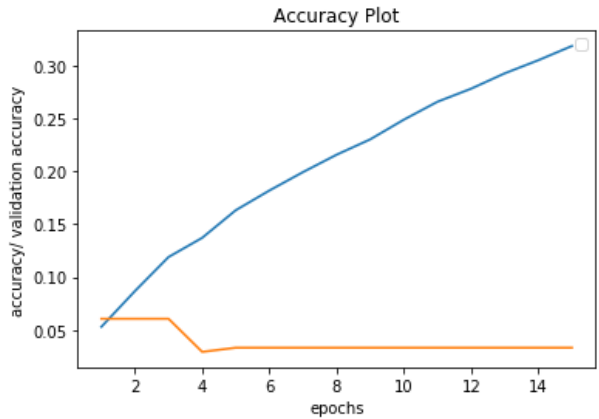

**Figure 12.** ResNet50 Accuracy Plot

[19] remarks that utilizing pre-trained models for transfer learning depends on the size of the dataset and its correlation to the features learned from the images on which the model has been trained. Furthermore, our dataset did not seem to be correlated to the objects and features used to train ResNet [20]. We hypothesize that ResNet performed poorly on our given dataset because the dataset is too small and ResNet's transferred features are not relevant.

To test this theory, we ran ResNet on the CIFAR dataset, with a learning rate of 5e-4,, for 200 epochs, using the categorical cross-entropy loss function, and Adam optimizer [21]. The final test accuracy resulted to 91.9%.

*4) InceptionV3*

The best Inception model used 2 additional hidden layers of 1536 Nodes, with dropouts of 0.5, leaning rates of .00005 and the ADAM optimizer using categorical cross entropy loss functions. Training was halted at 21 epochs.

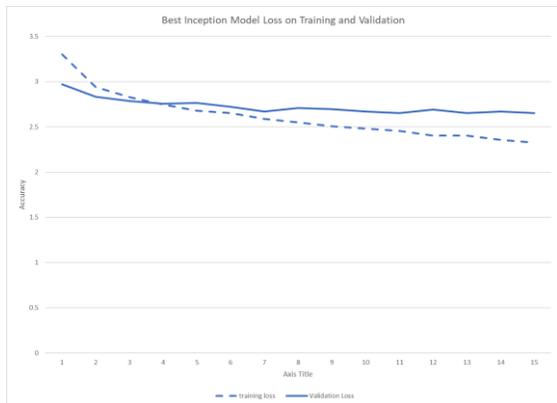

**Figure 13.** InceptionV3 Loss Plot

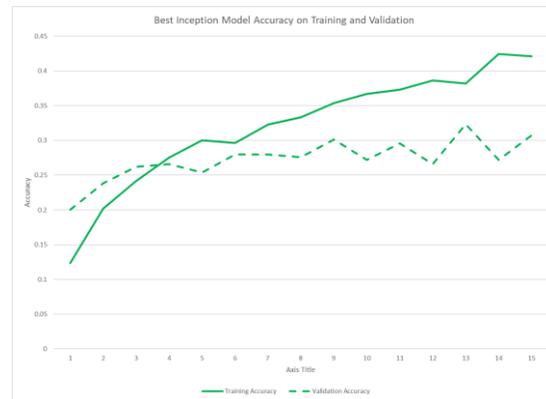

**Figure 14.** InceptionV3 Accuracy Plot

Inception and VGG19 performed better than VGG16 and ResNet. Inception's base classifier is trained on over 10 million images, many of which often included different orientations and partial obstruction, much like our bee data set contains; this may account for its comparatively higher performance with a little fine tuning.

*B. Composite Model*

We tried various combinations of composite models (a summing of each model's softmax output then select the highest category (ies)) and found that the composite model outperformed the best individual model.

**Table 4.** Best Model Performances

|  | Single Class Acc | Top-3 Acc |
|---|---|---|
| VGG19 | 19.7% | 40.2% |
| VGG16 | 15.7% | 39% |
| RESNET 50 | 0.0% | 7.4% |
| InceptionV3 | 23.6% | 44.5% |
| Inc + VGG19 | 25.5% | 50.3% |
| Inc + VGG 19 + VGG 16 | 27.5% | 50.4% |
| All combined | 25.5% | 40.6% |

*C. Confusion Matrix*

The confusion matrix from our best composite model is located in the Appendix. We note that only two (0.3%) bumble bees were mischaracterized as honey bees. Additionally, recall and precision were poor, see Appendix.

VII. CONCLUSION AND FUTURE WORK

Based on the image set we did not get great accuracy for either single species or top-3 classification from individual or composite models. While we achieved eight-times better accuracy than sheer guessing, this is probably not better than a skilled amateur can accomplish.

We found Inception performed better than VGG, and that ResNet is not well suited for this particular transfer learning task. This is not a statement of ResNet's suitability for all transfer learning, only that it was unsuited for our small image set of bumble bees.

Our results at species classification, based on natural bee images, are significantly worse than DrivenData's results from the same type of images. However, the observable differences between genus (honey vs bumble) may not be as difficult a problem as detecting more subtle intra-genus species differences. "Honeybees have a clear distinction between head and abdomen, bumblebees are 'all of one piece.' Honeybees also have two clear sets of wings: a larger set in front and a smaller set in back [16]". Notably, we have a very low rate of misidentifying bumble bees as honey bees.

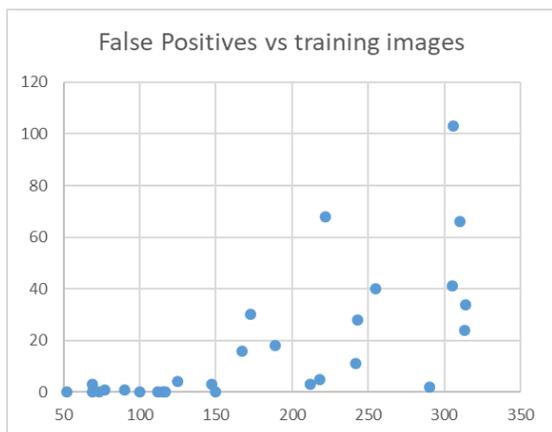

**Figure 15.** Differences between Honey and Bumble Bees[15]

### A. Need for More Data

Our first desire would be to acquire many more (an order of magnitude more) labeled images. Our models begin rapidly overtraining indicating there is not enough variation to present a large learning challenge. Our data was reasonable distributed, but at best a class was represented by 351 images and at worst 59. We feel this encourages the models to try and memorize the training data.

Next, we would look for unobstructed images of bees. A common entomological practice is to pin and mount insects for display and study. If we could source a large number of profiles, top and front images of previously mounted and identified bumble bees, it could aid learning species differences.

We observed that the composite model has a pronounced tendency to mistakenly categorize images as those where it had larger training sets, see Figure 16.

**Figure 16.** False Positives vs Training Image Number

We note that few false positives occur when the training base size was less than 150 images. We suspect the classifiers did not learn generalizable features for these species and hypothesize that the misclassifications would be more randomly distributed if all species had over 150 images.

All of the "zero" values for precision and recall, caused by a lack of true positive classifications, came from species with training data sets below 150 images, see figures in the Appendix.

### B. A Top-3 Loss Function

Then we could investigate or build a custom "top three" loss function. We feel this may achieve better results than cobbling together the top-3 from a strict summation of the individual classifier's softmax activations.

### C. Different Type of Composite Model

Finally, we envision a different type of composite model based on the proposition that different pretrained models have different strengths at identifying the important species differentiating features. We would build an encoder from the trained models, and then feed their concatenated outputs into a new neural network which can then train based on the learned features of the classifier-based encoder model.

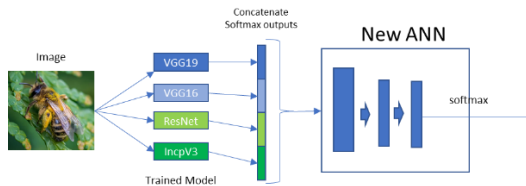

**Figure 17.** Alternate Composite Model

APPENDIX

**Figure 18.** Best Composite (VGG19, VGG16, Inception) Model Confusion Matrix

**Table 5.** Recall and Precision by Species

|  | Recall | Precision |
|---|---|---|
| Affinis | 0.0% | 0.0% |
| Appositus | 0.0% | 0.0% |
| Auricornus | 38.9% | 15.2% |
| Bifarius | 43.8% | 12.1% |
| Bimaculatus | 15.6% | 4.7% |
| Borealis | 7.1% | 16.7% |
| Californicus | 0.0% | 0.0% |
| Centralis | 0.0% | 0.0% |
| Citrinus | 12.5% | 50.0% |
| Fernaldae | 0.0% | 0.0% |
| Fervidus | 44.4% | 6.4% |
| Flavifrons | 0.0% | 0.0% |
| Fraternus | 53.8% | 31.8% |
| Griseocolis | 65.0% | 11.5% |
| Huntii | 18.8% | 18.8% |
| Impatiens | 18.5% | 20.8% |
| Insularis | 0.0% | 0.0% |
| Melanopygus | 41.9% | 5.5% |
| Mixtus | 19.2% | 22.7% |
| Nevadensis | 50.0% | 18.1% |
| Occidentalis | 0.0% | 0.0% |
| Pensylvanicus | 48.1% | 21.6% |
| Perplexus | 50.0% | 16.3% |
| Rufocinctus | 33.3% | 19.2% |
| Sonorus | 28.6% | 18.8% |
| Ternarius | 21.7% | 5.4% |
| Terricola | 0.0% | 25.0% |
| Vagans | 0.0% | 0.0% |
| Vosnesenskii | 11.1% | 16.7% |

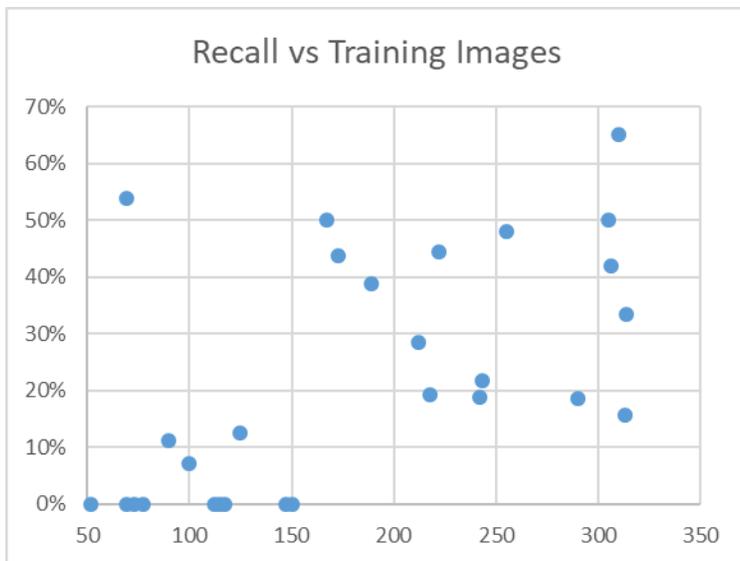

**Figure 19.** Recall vs Training Image Number

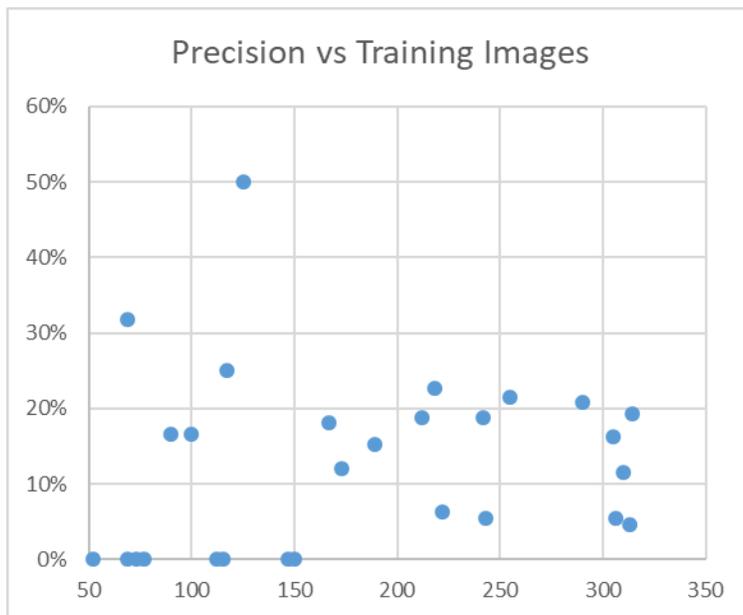

**Figure 20.** Precision vs Training Image Number